\documentclass[letterpaper, 10 pt, conference]{ieeeconf}  

\IEEEoverridecommandlockouts                              

\usepackage{cite}
\usepackage{amsmath,amssymb,amsfonts}
\usepackage{algorithmic}
\usepackage{graphicx}
\usepackage{textcomp}
\usepackage{xcolor}
\usepackage{tabularx}
\usepackage{threeparttable}
\usepackage{multirow}
\usepackage{multirow}
\usepackage{array}
\usepackage{booktabs}
\usepackage{siunitx}
\usepackage{makecell}
\usepackage{hyperref}
\hypersetup{%
 hidelinks,
 colorlinks=false}

\title{\LARGE \bf
Layout-independent actuation allocator for fin-actuated marine robots
}

\author{Yuya Hamamatsu$^{1}$, Maarja Kruusmaa$^{1}$, Asko Ristolainen$^{1}$
\thanks{$^{1}$Department of Computer Systems, Tallinn University of Technology, Tallinn, Estonia
        {\tt\small (Yuya.Hamamatsu, Maarja.Kruusmaa, Asko.Ristolainen)@taltech.ee}, 
        }
}

\begin{document}

\maketitle
\thispagestyle{empty}
\pagestyle{empty}

\begin{abstract}
In this study, we propose a layout-independent control allocator capable of zero-shot deployment across diverse actuator configurations. The proposed method utilizes a learning pipeline that integrates a Graph Neural Network (GNN) and a Transformer to represent the robot's geometric layout as a graph, along with a Mixture Density Network (MDN) to predict multi-modal control command distributions. Furthermore, by incorporating a differentiable physics surrogate model, we achieve command refinement during inference to minimize target wrench tracking error and energy consumption. A single generalized model using randomly generated actuator layout data demonstrated high trajectory tracking performance on different actuator layout robots outside the training distribution. Additionally, in real-world pool experiments, our approach achieved performance nearly equivalent to conventional controllers designed to specific layouts.
\end{abstract}

\section{Introduction}
In modern marine robotic systems, one of the key challenges is mapping the desired body-frame force and torque (wrench) to the appropriate commands for each actuator \cite{johansen2013control}. For a more optimal allocation, it is necessary to use an advanced inverse model to design a combination of multiple commands that output the same wrench. This depends on the configuration of the actuators in the robot and must be designed individually. If the layout changes or some actuators fail, the controller needs to be re-designed or re-tuned for the new configuration. This dependence on a predefined layout makes it difficult to quickly adapt to design changes or handle actuator faults. 

For ground-based robots, such as robot arms, several learning-based methods have been proposed to deal with this diversity of robot layouts \cite{10611477}, \cite{Doshi24-crossformer}. However, these methods do not cover actuators for underwater robot navigation. In addition, depending on the actuator configuration of an underwater robot, multiple command combinations are required to generate the target force. In such cases, it is necessary to select a command combination from among the multiple diverse combinations, taking into account energy efficiency and stability. Underwater fin actuation yields a highly nonlinear and often nonconvex allocation problem. Therefore, optimization-based methods can suffer from local minima or singular regions induced by constraints unless carefully formulated \cite{johansen2008optimal}, \cite{berge1997robust}. This problem is more substantial when a single actuator, such as a fin, can generate forces in multiple directions. Previous research has shown that optimizing this assignment problem has significantly improved energy efficiency \cite{fossen2006survey}, \cite{REMMAS2026106646}. Research using reinforcement learning has covered multiple layouts for underwater robots, but this research requires transfer learning, which requires additional learning to adapt the policy to specific robots \cite{hamamatsu2025icra}.

\begin{figure}[t]
\includegraphics[width=1.00\linewidth]{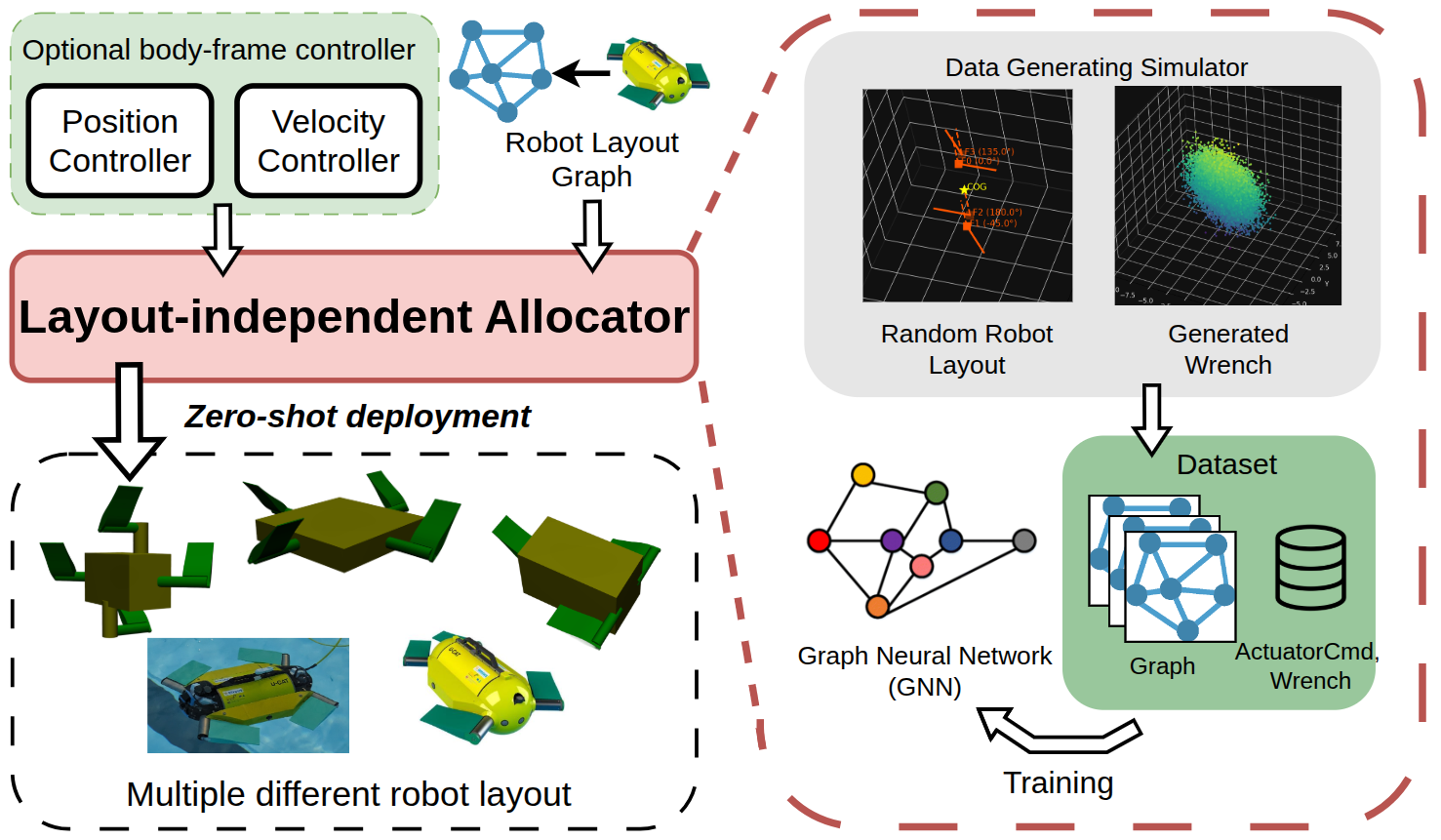}
\caption{Overview of the layout-independent control allocator for marine robots.}
\label{overview}
\end{figure}

In this work, we propose a layout-independent controller that can be generalized across different actuator configurations. Our proposed pipeline overview is shown on Fig. \ref{overview}. Our approach uses a Graph Neural Network (GNN) to represent the robot’s layout, such as the positions and orientations of the fins, as a graph and learns to predict the control commands necessary to achieve a target wrench \cite{wu2020comprehensive}, \cite{velivckovic2017graph}. We generate a large training data set in simulation by randomly selecting the configuration of the robot layout and applying a random command input to record the resulting forces and torques in the robot’s body frame. Using these data, the proposed GNN-based pipeline learns the relationship between the geometrical layout of the robot and the mapping of commands to the output body-frame wrench. Furthermore, the proposed pipeline includes a surrogate neural network that approximates the physical model used for inference refinement. We exploit the differentiability of the surrogate model to optimize low energy actuator commands while matching the target body-frame wrench \cite{sanchezgonzalez2018graphnetworkslearnablephysics}, \cite{newbury2024review}. Once trained, our pipeline can be applied to new robot layouts  using only their actuator geometrical information to compute the commands for a desired wrench without any additional training. This layout information consists only of positions and angles of the layout and does not include parameters requiring complex system identification. The same trained pipeline serves as a generalized controller that is not tied to a specific robot layout, achieving zero-shot generalization to different layouts. 

\begin{figure*}[t]
\includegraphics[clip, width=17.7cm]{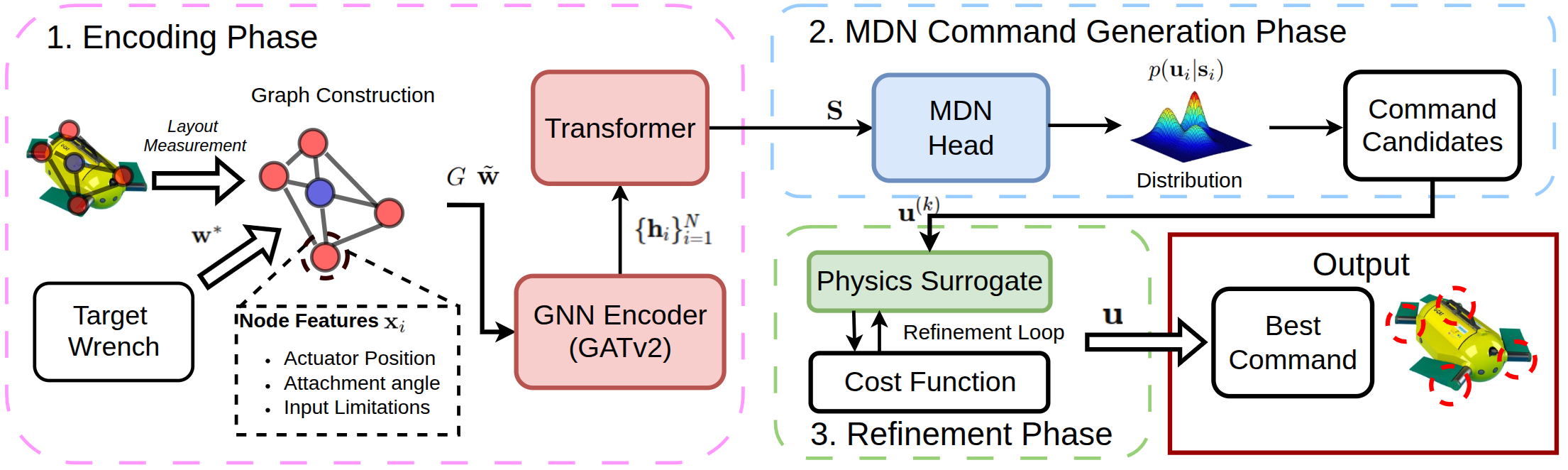}
\caption{Pipeline of the proposed method. The system takes a target wrench and robot layout graph as input. The Graph Encoder (GATv2) and Transformer extract features to predict a mixture of distributions via the MDN Head. $K$ candidate control commands are sampled and passed to a differentiable Physics Prediction Head. The candidates are ranked and refined via gradient descent optimization to minimize a composite cost function comprising task error, energy consumption. }
\label{pipeline}
\end{figure*}

We demonstrate the performance of the proposed pipeline in many different layouts of a fin-actuated underwater robot, both in simulations and in real-world experiments. A single trained pipeline is used to allocate control commands for robots with multiple layouts and consistently produces the desired forces and torques in the body frame in these variations. We also evaluate scenarios in which one or more fins experience failure, to test the controller’s robustness in the case of actuator loss. The results show that our layout-agnostic controller can successfully maintain control performance even under partial actuator failures. This experiment was carried out in the real world to evaluate sim2real performance and the impact of asymmetric and different fin counts. This approach contributes to more flexible and resilient control systems that can handle changes in robot layout or unforeseen component failures.

The main contribution of our method is that it demonstrates the feasibility of a generalized zero-shot allocation controller that can handle multiple robot layouts. Our proposed method can also handle actuator failures without a pre-designed fault-tolerant controller.

\section{Layout-independent allocation}

We propose control allocation from a desired body-frame wrench $\mathbf{w}^\ast\in\mathbb{R}^6$ to per-fin commands $\mathbf{u}_i=[A_i,f_i,\phi^{c}]\in\mathbb{R}^3$, where $A$ is amplitude, $f$ is frequency, and $\phi^{c}$ is the zero-direction angle. The possible allocation combination is generally not unique. In addition, the number and placement of the fins may change between robots. Our goal is to have a layout-independent allocator that uses only the layout description at inference time. We represent the layout as a graph and train a network that outputs a distribution over feasible per-fin commands conditioned on the layout. The allocator uses a GNN \cite{velivckovic2017graph}, a Transformer \cite{vaswani2023attentionneed}, and a Mixture Density Network (MDN) \cite{bishop1994mixture}. We use a GNN to learn how the fins layout geometry and connectivity shape control-allocation patterns, producing layout-aware node embeddings that generalize across varying fin counts and placements, and we use MDN to model the multi-modal inverse mapping by outputting a distribution over per-fin commands for the same target wrench. This design allows to select the most optimal combination of commands that can output the same wrench. We train an Multi Layer Perceptron (MLP) surrogate as a differentiable forward model that predicts the wrench from candidate commands. Starting from MDN samples, we backpropagate a cost function to refine the commands and select the best candidate.
This physics surrogate is used during training as guidance and during inference for candidate selection and refinement to select optimal command allocation at inference. Each robot uses multiple independently controlled silicone fins ($i$) that generate thrust through flapping motion, regulated by amplitude $A_i$, zero-direction $\phi^{c}_i$, and frequency $f_{i}$. The oscillatory movement for each fin at time $t$ is described by:

\begin{equation}
\label{osc}
\theta(t)=A_{i} \sin \left(2\pi f_{i} t\right)+\phi^{c}_i
\end{equation}

\noindent where $\theta(t)$ is set as the next target angle of the fin \cite{remmas2021inverse}. These $(A_i,f_i,\phi^{c}_i)$ are used as input commands for each fin in the following sections. The pipeline diagram is shown in Fig. \ref{pipeline}.

\subsection{Graph representation of robot layout}

We represent each robot layout as a fully connected undirected fin graph $G=(V,E)$ to simply support a variable number of fins. Each node $i$ represents a fin with features $\mathbf{x}_i \in \mathbb{R}^{10}$. These comprise the position of the fin relative to the center of gravity $\mathbf{r}_i \in \mathbb{R}^3$, its mounting angle $\alpha_i$, and six variables that define the operational limits of min/max for amplitude, frequency, and zero-direction.
We normalize the target wrench using dataset statistics with mean $\boldsymbol{\mu}$ and standard deviation $\boldsymbol{\sigma}$ as $\mathbf{\tilde{w}} = \text{diag}(\sigma)^{-1} (\mathbf{w}^* - \mu)$.


\begin{table*}[t]
\centering
\caption{Ablation of the inference pipeline}
\label{tab:ablation_pipeline}
\small
\setlength{\tabcolsep}{4pt}
\begin{tabular}{lccc}
\toprule
Variant & Normalized Error & Wrench Error & Time [s] \\
\midrule
Mean only ($\bar{u}$)                                  & \num{3.64e-02} $\pm$ \num{4.06e-02} & \num{4.25e-02} & 0.0078 \\
without refinement                                    & \num{1.01e-01} $\pm$ \num{1.59e-01} & \num{1.51e-01} & 0.0028 \\
MDN NLL selection + refinement                          & \num{3.76e-03} $\pm$ \num{4.84e-03} & \num{6.89e-03} & 0.0801 \\
Random warm-start + refinement                          & \num{2.28e-01} $\pm$ \num{3.88e-01} & \num{2.11e-01} & 0.0772 \\
Ours (MDN + Physics-guided + refinement)      & \num{3.03e-04} $\pm$ \num{2.65e-04} & \num{6.14e-04} & 0.0826 \\
\bottomrule
\end{tabular}
\end{table*}

\subsection{Allocation pipeline}
\subsubsection{Encoding Phase}
The input is the fin graph $G$ and the normalized target wrench $\tilde{\mathbf{w}}$.
We first project the node feature of each fin $(\mathbf{x}_i)$ embedding using a shared linear projection.
We then apply a GATv2 layer \cite{velivckovic2017graph} that updates each node embedding by aggregating information from other fins, where the contribution of each neighbor is weighted by learned attention scores. To condition the allocator on the desired motion, we pass the normalized target wrench $\tilde{\mathbf{w}}$ through a separate linear projection to obtain a wrench embedding $\mathbf{c}$. We replicate $\mathbf{c}$ across all fin nodes in the same graph and concatenate it into each node embedding. After the GATv2 update and the conditioning of the wrench, we denote the resulting node embeddings by $\mathbf{h}_i$, which are conditioned on both the fin layout and the target wrench. Next, we apply a Transformer encoder over the set of fin embeddings ${h_i}_{i=1}^{N}$ so that each fin can attend to all others and coordinate a command set that satisfies the global wrench. Let $\mathbf{S}=\mathrm{Transformer}([h_1,\dots,h_N])$ and denote its i-th output by $\mathbf{s}_i$. For minibatches with varying (N), we use standard padding and a key-padding mask so that nonexistent fins do not participate in self-attention.

\subsubsection{MDN Command Generation Phase}
From $\mathbf{s}_i$, an MDN predicts a mixture of Gaussians on $\mathbf{u}_i$.
With mixture components $M$, the MDN outputs mixture weights $\pi_{i,k}$,
means $\boldsymbol{\mu}_{i,k}\in\mathbb{R}^3$,
and diagonal standard deviations $\boldsymbol{\sigma}_{i,k}\in\mathbb{R}^3$:
\begin{equation}
p(\mathbf{u}_i|\mathbf{s}_i)=\sum_{k=1}^{M}\pi_{i,k}\,
\mathcal{N}(\mathbf{u}_i;\boldsymbol{\mu}_{i,k},\mathrm{diag}(\boldsymbol{\sigma}_{i,k}^2)).
\end{equation}
This formulation explicitly represents multi-modality in allocation.
We can compute the expected mixture command 
$\bar{\mathbf{u}}_i=\sum_k \pi_{i,k}\boldsymbol{\mu}_{i,k}$. To control the exploration–exploitation trade-off during candidate generation, we temper the MDN mixture weights with a sampling temperature $T$:
\begin{equation}
\pi^{(T)}_{i,k}=\mathrm{softmax}\!\left(\frac{\log \pi_{i,k}}{T}\right).
\end{equation}
We then sample the component index $k\sim\mathrm{Cat}(\boldsymbol{\pi}^{(T)}_i)$ and draw $\mathbf{u}_i$ from the corresponding Gaussian.

\subsubsection{Refinement Phase}
We integrate a physics surrogate into the pipeline.
The physics surrogate is differentiable as
$\hat{\mathbf{w}}=\phi(\mathbf{u},G)$
that reconstructs the body-frame wrench from commands and layout.
This provides a fast, layout-conditioned, and differentiable evaluation of candidate allocations, allowing physics-guided selection among multiple MDN samples and gradient-based refinement of commands to reduce wrench error and energy without requiring iterative calls to a high-fidelity simulator.

In our implementation, the head predicts the per-fin force with an MLP and computes the torque analytically.
Let $\mathbf{F}_i\in\mathbb{R}^3$ be the predicted force for fin $i$:
\begin{equation}
\mathbf{F}_i = f_{\psi}([\mathbf{x}_i,\mathbf{u}_i]),
\qquad
\boldsymbol{\tau}_i = \mathbf{r}_i \times \mathbf{F}_i,
\qquad
\hat{\mathbf{w}} = \sum_i [\mathbf{F}_i;\boldsymbol{\tau}_i].
\end{equation}
The physics surrogate also provides per-fin wrench contributions, which can be used for interpretability and failure analysis. It computes a global layout descriptor by pooling node-wise layout embeddings and produces scale and bias terms for a 6D affine correction.

\begin{figure*}[t]
\includegraphics[width=\linewidth]{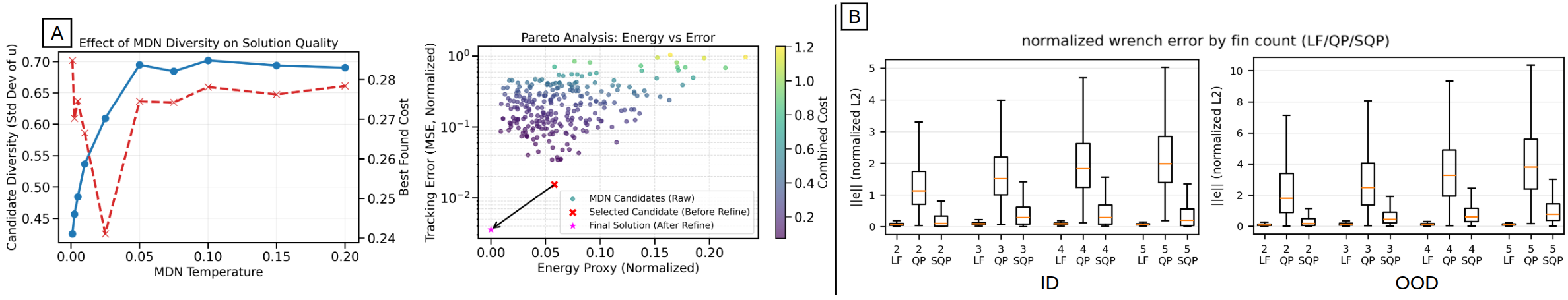}
\caption{A. (Left) Effect of MDN sampling temperature and refinement. (Left) Candidate command diversity and the best selection cost J as a function of the MDN temperature T. (Right) Distribution of raw MDN candidates in the energy-proxy vs. wrench error. B. 100 Layout analysis results. In-Distribution(ID)/Out-Of-Distribution(OOD) wrench error grouped by fin count (2-5 fins), }
\label{ab}
\end{figure*}

\subsection{Data-collection and training process}

\subsubsection{Data generation}

We generate training data by forward sampling. The training dataset contains uniformly sampled absolute actuator commands $u$ within the actuator limitation and calculate the resulting body-frame wrench $w_{\mathrm{body}}$. The simulated actuation dynamics integrates the body-frame wrench contributions of the fins. This procedure produces paired samples $(u, w_{\mathrm{body}})$ to learn an inverse allocator into the proposed pipeline. We use $(u, w_{\mathrm{body}})$ as supervision, noting that the induced distribution on $w_{\mathrm{body}}$ is determined by the uniform sampling in $u$. 

\subsubsection{Physics surrogate pretraining}
We pretrain the physics surrogate separately from the main allocator to use this model as guidance during main training.
At each iteration, we randomly choose one layout and sample a batch of commands.
We compute the true wrench using an analytic fin model and fit the surrogate to match it.
The main loss is the MSE on the global wrench in normalized space:
\begin{equation}
\mathcal{L}_{phys} = \| (\hat{w} - \mu) \oslash \sigma - (w - \mu) \oslash \sigma \|_2^2
\end{equation}

\subsubsection{Main training with physics surrogate guidance}
We train the allocator network (GNN + Transformer + MDN) while keeping the physics surrogate weights fix.
The objective consists of an imitation loss, a physics-guidance loss computed through $\phi$, and an entropy regularizer on the mixture weights to prevent component collapse.
The imitation loss is the negative log-likelihood of the MDN of the recorded commands $\mathbf{u}^\star$.

For physics guidance, we pick up samples $\mathbf{u}^{(s)} \sim p_{\mathrm{MDN}}(\mathbf{u}\mid G,\tilde{\mathbf{w}})$ from the predicted mixture distribution and evaluate the resulting wrench using the frozen physics head.
We compare the predicted wrench to the target wrench in the normalized wrench space:
\begin{equation}
\mathcal{L}_{\mathrm{guide}}
=
\bigl\|\ \mathcal{N}\!\left(\phi(\mathbf{u}^{(s)},G)\right)\ -\ \tilde{\mathbf{w}}\ \bigr\|_2^2 ,
\end{equation}
where $\mathcal{N}(\cdot)$ applies the same normalization as used to obtain $\tilde{\mathbf{w}}$.
Using samples penalizes implausible modes and encourages the entire predicted distribution to place probability mass on allocations that are physically consistent.
For stability, we linearly ramp the guidance weight $\beta$ from a small value to its final value. To discourage the mixture from collapsing to a single component, we maximize the entropy of the mixture weights $\boldsymbol{\pi}$. 
The final objective is
\begin{equation}
\mathcal{L}
=
\mathcal{L}_{\mathrm{imit}}
+
\beta\,\mathcal{L}_{\mathrm{guide}}
+
\lambda\,\mathcal{L}_{\mathrm{ent}}.
\end{equation}
where $L_{\mathrm{imit}}=-\log p_\theta(u^{*}\!\mid G)$ and $L_{\mathrm{ent}}:=\sum_{k=1}^{K}\pi_k\log\pi_k=-H(\boldsymbol{\pi})$ with entropy regularizer on mixture weights $\boldsymbol{\pi}$.

\subsection{Inference}

At inference time, we only need the layout configuration and the desired wrench $\mathbf{w}^\ast$. We deterministically construct the fully-connected fin graph $G$ from the measured fin geometry (positions/orientations) and per-fin limits, and normalize the wrench to $\tilde{\mathbf{w}}$ using the same statistics. We compute MDN parameters and sample multiple candidate allocations from the MDN. Each candidate is clamped to the per-fin limits stored in the node features. For each candidate command of all fins $k$, we evaluate the predicted wrench $\hat{\mathbf{w}}^{(k)}$ using the physics surrogate.
We batch all candidates and compute them with a single forward pass for efficiency.
We score each candidate with the normalized tracking error:
\begin{equation}
J^{(k)}_{\mathrm{norm}}
=
\|\;(\hat{\mathbf{w}}^{(k)}-\boldsymbol{\mu})\oslash\boldsymbol{\sigma}
-
\tilde{\mathbf{w}}\;\|_2^2 .
\end{equation}
The selected candidates are ranked with the cost function score for refinement.
We normalize each command by its limit and penalize the squared magnitude:


\begin{equation}
J_{\mathrm{en}}^{(k)}=\sum_{i=1}^{N}\!\left(
w_A A_{n,i}^2+w_f f_{n,i}^2+w_\phi \phi_{n,i}^2+w_{Af}A_{n,i}f_{n,i}
\right),
\end{equation}

$A_{n,i}$, $f_{n,i}$, and $\phi_{n,i}$ are the amplitude, frequency, and zero-direction of fin $i$, each normalized by its corresponding upper limit of the dataset. $w_A$, $w_f$, $w_\phi$, and $w_{Af}$ are weighting coefficients. The final selection score is $J^{(k)} = J^{(k)}_{\mathrm{norm}} + \gamma J_{\mathrm{en}}$.   After scoring the $K$ MDN-sampled candidates with the physics head, we take the top $M$ ranked candidates and refine them by directly optimizing the command variables to reduce the tracking the cost function $J^{(k)}$. Gradients with respect to the commands are obtained via backpropagation through the physics head, and the commands are updated using Adam~\cite{kingma2014adam}. 

\begin{table*}
\caption{100 robot layouts wrench tracking error}
\label{tab:wrench_overall_split_method}
\begin{tabular}{llrrrrrrrr}
\toprule
split & method & n & err norm mean & err norm med & err norm p95 & err norm max & err phys mean & success@0.5 & success@1.0 \\
\midrule
ID & Ours & 100000 & 0.0857 & 0.0769 & 0.1726 & 1.1244 & 0.1019 & 0.9997 & 1.0000 \\
ID & QP & 100000 & 1.8410 & 1.6072 & 3.9688 & 13.2264 & 2.2520 & 0.0488 & 0.2324 \\
ID & SQP & 100000 & 0.3983 & 0.2110 & 1.4476 & 7.8580 & 0.4914 & 0.7267 & 0.8920 \\
OOD & Ours & 100000 & 0.1157 & 0.0922 & 0.2831 & 5.9936 & 0.1395 & 0.9925 & 0.9995 \\
OOD & QP & 100000 & 3.3913 & 2.9457 & 7.7522 & 25.2299 & 3.4824 & 0.0358 & 0.1281 \\
OOD & SQP & 100000 & 0.8148 & 0.4904 & 2.7619 & 17.8511 & 0.8871 & 0.5069 & 0.7344 \\
\bottomrule
\end{tabular}
\end{table*}

\subsection{Implementation details}
This study generated 1000 robot layouts and calculated 5000 different commands for each robot to create a training dataset of 5000000 samples. We set fin layouts with 2–5 fins, sampling fin positions in the body frame within $(x,y\in[-0.30,0.30])$ m and $(z\in[-0.10,0.10])$ m, and mounting angles in $([-180,180])$ deg. Each fin is modeled as a node with 10-D features (Fins attachment position, center of gravity, mounting angle, and mechanical actuation limits), and we use a fully-connected undirected fin graph. The network consists of a 2-layer GATv2 encoder (4 heads, 128-dim), a 3-layer 4 heads Transformer encoder. An MDN head with $M{=}8$ components for 3-D fin commands $[A,f,\phi^{c}]$ with $\log\sigma\in[-3,3]$. The target wrenches are standardized using the mean/std in the data set (std floor $10^{-6}$). We train the model for 115 epochs with Adam (learning rate $0.0001$, batch 16384). The physics head predicts the per-fin forces with an MLP (hidden 512) and computes the torques. 

\subsection{Model analysis}
We analyze how temperature-controlled MDN sampling, Physics-guided candidate ranking, and gradient-based refinement jointly affect candidate diversity and final wrench-tracking accuracy.
As shown in Fig.~\ref{ab}A (left), diversity grows monotonically with $T$, but the best selection cost $J$ is minimized at an intermediate temperature:
too small $T$ under-explores alternative solutions, while too large $T$ yields many diverse yet low-quality candidates.
Fig.~\ref{ab}A (right) further illustrates the effect of refinement: starting from a selected candidate, gradient-based refinement consistently reduces wrench-tracking error while remaining in a low-energy region. 

Table~\ref{tab:ablation_pipeline} evaluates the pipeline under a controlled protocol: all variants use the same test set, the same number of candidates, and the same temperature $T$; when refinement is enabled, identical refinement hyperparameters are used.
Each row modifies exactly one stage (initialization, candidate ranking, or refinement).
Removing refinement degrades performance, indicating that local optimization is required for high-precision wrench matching.
Replacing Physics surrogate-guided ranking with NLL selection that choosing the sampled candidate with the highest MDN likelihood increases error by more than an order of magnitude, suggesting that the learned prior alone does not reliably rank candidates for the true control objective; Physics-guided evaluation is therefore important for selecting a good warm start.
Finally, the performance of a random feasible warm start followed by refinement is decreased, confirming strong non-convexity and showing that the MDN prior provides a necessary initialization that places refinement in a favorable basin of attraction.

\subsection{Baseline methods}
We compare against two model-based optimizers, Quadratic Programming (QP) and Sequential Quadratic Programming (SQP) under the same per-fin box constraints on $(A,f,\phi^c)$.
QP precomputes, for each layout, a first-order approximation of the actuation map by central finite differences using the analytical actuation map and then solves a regularized box-constrained least-squares problem with an active-set solver. SQP iteratively re-linearizes the map at the current iterate, solves a box-QP on the normalized residuals, and applies a backtracking line search; Jacobians are obtained via proxy autograd by default, or via finite differences of the analytical map.

In the real-world experiments, we use a layout-specific analytic allocator as a baseline to compare the proposed layout-independent method. We use Remmas et al. analytic control allocation as our baseline, which has been reported to outperform optimization-based allocation (QP/SQP) in simulation and pool experiments, while being significantly more computationally efficient \cite{REMMAS2026106646}. We also use this method as a fault-tolerant baseline for the fin failure scenario \cite{remmas2023fault}. 

\section{Body-frame control integration}
The previous section presented a generic layout-independent allocation framework that takes a desired body-frame wrench as input. Because the allocator is agnostic to how this wrench is generated, any body frame controller can be connected upstream as long as it outputs $(F_x, F_y, F_z, \tau_x, \tau_y, \tau_z)$ or the required subset in the body frame. In this study, we therefore treat the body-frame controller as a plug-in module and evaluate the framework using two controllers that were actually used in our experiments: a waypoint-based trajectory tracking controller that produces $(F_x, F_z, \tau_z)$ from odometry and a target waypoint, and a velocity-and-depth controller that produces $(F_x, F_z, \tau_z)$ from Doppler Velocity Log (DVL) and pressure sensor feedbacks. The following subsections describe these two high-level controllers and how their wrench commands are interfaced to the allocator. 

\subsection{Trajectory tracking controller}
This controller generates a desired body-frame wrench to track a waypoint \((x_t, y_t, z_t)\) using odometry \((x, y, z, \psi)\). Calculate the position error \((dx, dy, dz)\) and the planar distance. The desired heading is set toward the waypoint, and the yaw error is wrapped to the range \([-\pi,\pi]\). A proportional yaw torque command \(\tau_z = k_{\psi} e_{\psi}\) aligns the vehicle to face the target. The surge force command is proportional to the planar distance and is scaled by \(\max(0,\cos e_{\psi})\) so that forward thrust is reduced when the heading is misaligned and rotation is prioritized. The heave force command is proportional to the depth error \(F_z = k_z dz\). In our real-world experiment, the pose feedback was obtained using ArUco markers \cite{ar} and the camera mounted on the robot.

\subsection{Velocity controller}
The velocity control module generates a desired body-frame wrench to track a reference surge velocity and a reference depth using onboard sensors. The surge velocity is estimated from the DVL measurements after simple filtering, then a PID controller computes the surge force \((F_x)\) from the velocity error. The depth is calculated from the fluid pressure, and a PID controller computes the heave force \((F_z)\) from the depth error. The yaw motion is damped using a proportional controller to the yaw rate of the IMU, producing a yaw torque \((\tau_z)\). The final command is published as a wrench \((F_x, F_z, \tau_z)\) with safety saturation and is used as input to the same downstream allocation stage. 

\begin{table}[t]
\centering
\footnotesize
\setlength{\tabcolsep}{5pt}
\renewcommand{\arraystretch}{1.10}
\caption{Robot layout configuration}
\label{tab:layout_abcd}

\begin{tabular}{c c
                S[table-format=+1.3]
                S[table-format=+1.3]
                S[table-format=+1.3]
                S[table-format=+3.0,round-precision=0]}
\toprule
Layout & Fin $i$ & {$x_i$} & {$y_i$} & {$z_i$} & {$\alpha_i$ [deg]} \\
\midrule

\multirow{4}{*}{A} &
 1 &  0.1502 &  0.25753 & 0.0 &   30 \\
 & 2 & -0.1502 &  0.25753 & 0.0 &  150 \\
 & 3 & -0.1502 & -0.25753 & 0.0 & -150 \\
 & 4 &  0.1502 & -0.25753 & 0.0 &  -30 \\
\midrule

\multirow{4}{*}{B} & 
 1 &  0.250 &  0.000 & 0.0 &    0 \\
 & 2 &  0.000 & -0.250 & 0.0 &  -90 \\
 & 3 & -0.250 &  0.000 & 0.0 &  180 \\
 & 4 &  0.000 &  0.250 & 0.0 &   90 \\
\midrule

\multirow{4}{*}{C} & 
 1 &  0.000 &  0.000 &  0.200 &    0 \\
 & 2 &  0.096 & -0.226 &  0.000 &  -45 \\
 & 3 &  0.000 &  0.000 & -0.200 &  180 \\
 & 4 & -0.096 &  0.226 &  0.000 &  135 \\
\midrule

\multirow{3}{*}{D} & 
 1 &  0.150 &  0.200 & 0.0 &    0 \\
 & 2 & -0.050 & -0.200 & 0.0 &    0 \\
 & 3 & -0.150 &  0.200 & 0.0 &  180 \\
\bottomrule
\end{tabular}
\end{table}

\begin{table}[t]
  \centering
  \caption{Different robot layout performance}
  \label{sim_result}
  \begin{tabular}{ccccc|c}
    
    Robot layout & A & B & C & D & All \\
    \hline
    ATE        & 0.1335 & 0.1435 & 0.1244 & 0.0937 & 0.1237 \\
    RTE        & 0.1265 & 0.1116 & 0.1755 & 0.0988 & 0.1281 \\
    Depth RMSE & 0.1084 & 0.1534 & 0.2023 & 0.0725 & 0.1341 \\ 
  \end{tabular}
\end{table}

\begin{figure*}[t]
\includegraphics[clip, width=17.7cm]{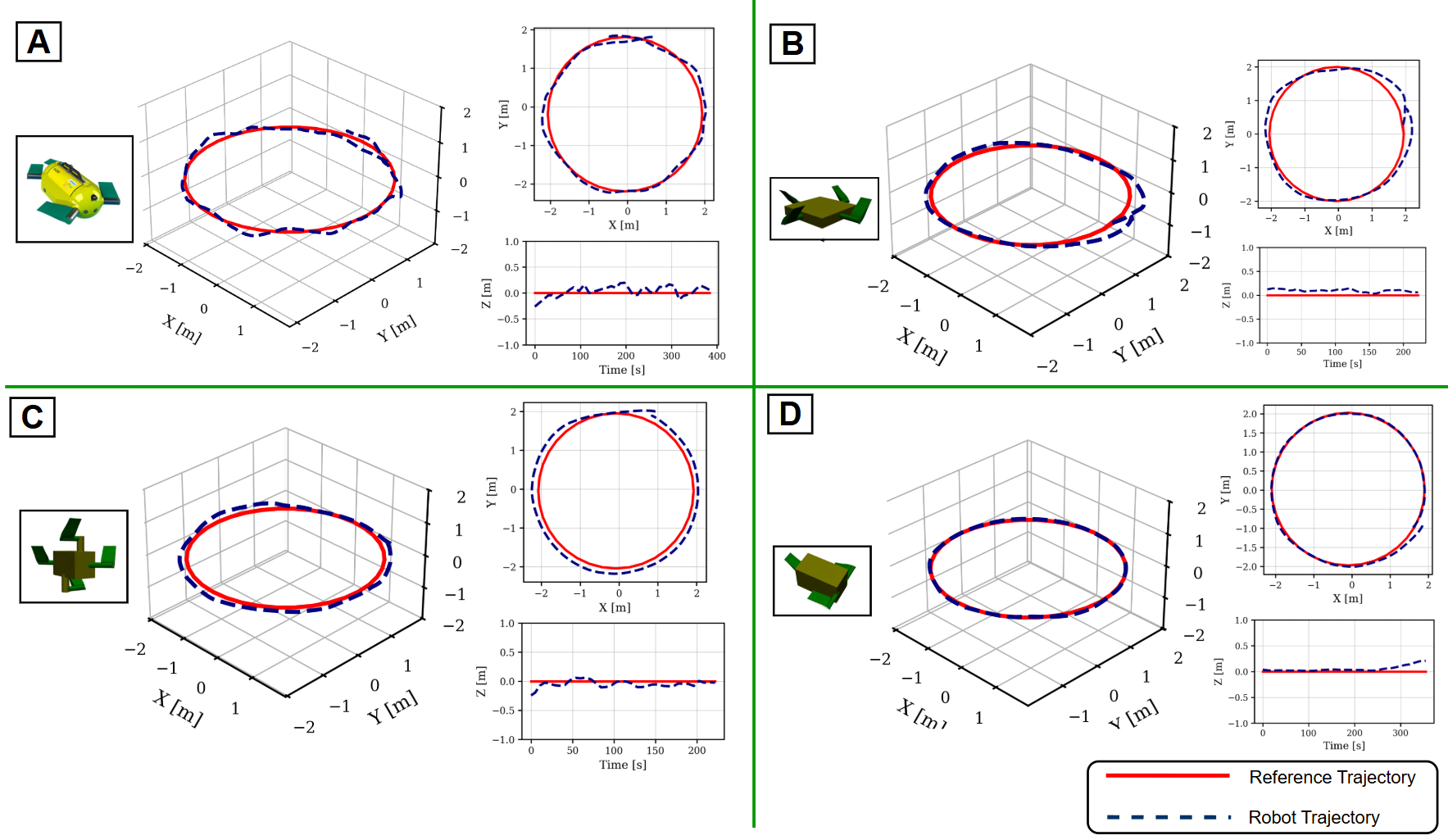}
\caption{Trajectory tracking results on the simulation using four different layouts. Each robot actuator configuration is describe in Table \ref{tab:layout_abcd}.}
\label{sim_fig}
\end{figure*}

\section{Experimental results}
To evaluate the performance of the proposed method, we conducted simulations and experiments. In the simulations, we verified the control performance of robots with different layouts. In real-world experiments, we evaluated the performance in a scenario where a fin failed and compared it with a baseline method. All closed-loop simulations and pool experiments were conducted on layouts intentionally omitted from the training. The same single model is deployed without any fine-tuning, using only the provided robot layout graph at inference. In all experiments, the initial MDN samples $K$ was set at 64, and the temperature $T$ was set at 0.02.

\subsection{100 different layout wrench tracking}

Firstly, we evaluate the allocator itself using a Monte Carlo setup. The layout group output within the robot distribution output during training is called the In-Distribution (ID) split, and the layout group output outside the training distribution is called the Out-Of-Distribution (OOD) split. These layouts are newly generated layouts that are not included in the training data. For each layout, we test 1000 target wrenches and compare the proposed layout-independent allocator (Ours) against a QP baseline under identical actuator limits. We also report \textit{success@}\,$\tau$, defined as the fraction of test cases whose normalized wrench-tracking error $\mathcal{E}_{norm}$ falls below a threshold $\tau$; therefore, \textit{success @ 0.5} and \textit{success @ 1.0} indicate the percentage of samples that achieve at most 50\% and 100\% of the reference error level, respectively. Table \ref{tab:wrench_overall_split_method} summarizes the results of 100 robot layout. In ID layout, our pipeline achieves the better performance in all metrics that QP/SQP. In OOD layouts, our method degrades moderately, but maintains the performance, while QP/SQP shows substantially larger errors and a lower success rate. These results indicate that the proposed allocator maintains reliable wrench tracking across out of training data distribution layouts, supporting subsequent closed-loop simulations. 

\begin{figure*}[t]
\includegraphics[clip, width=17.7cm]{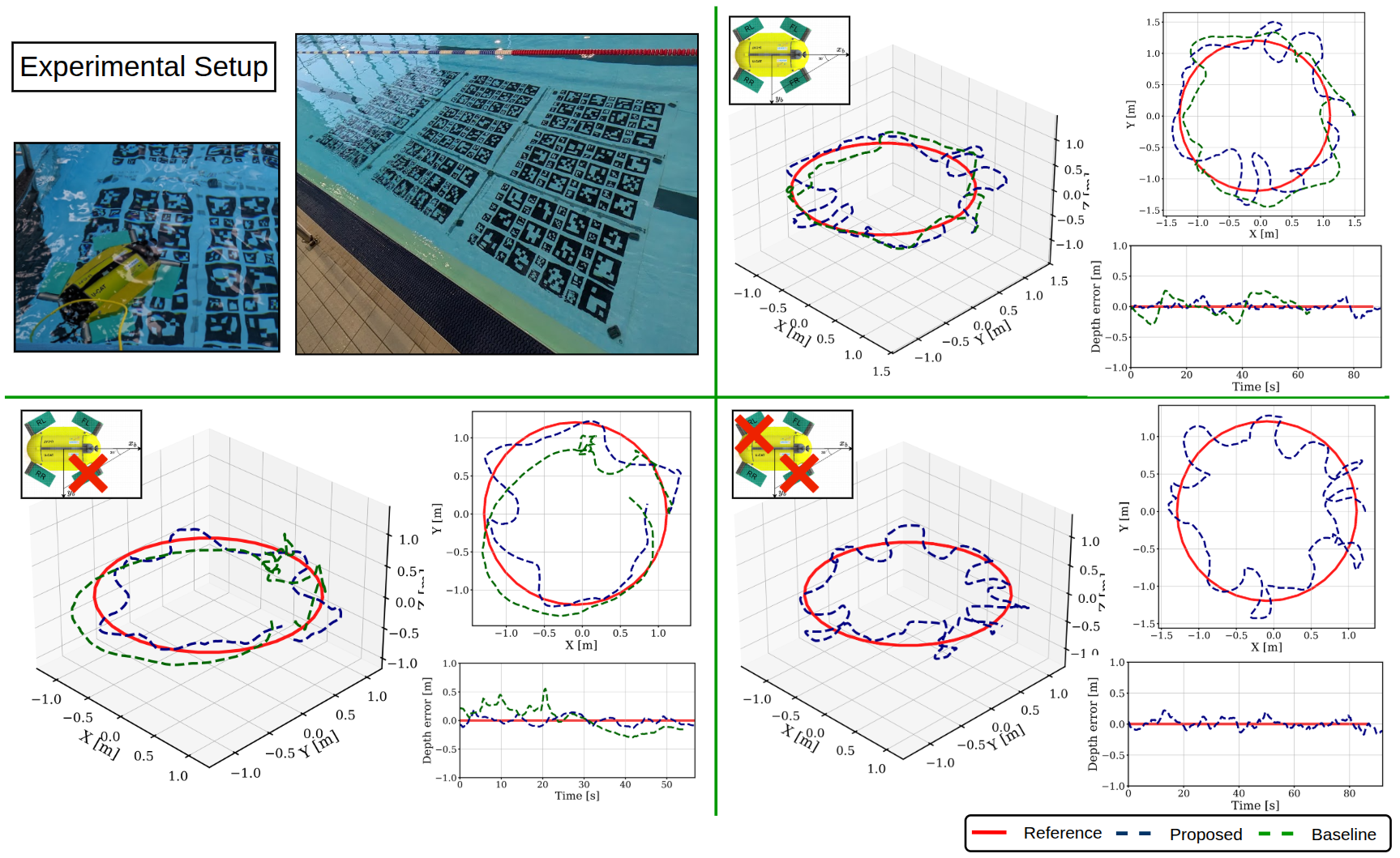}
\caption{Pool experiments for fault-tolerant tracking with fin failures. (Top-Left) Experimental setup. (Top-Right): tracking results for healthy, (Bottom-Left) one fin broken, and (Bottom-Right) two fins broken. Red: reference trajectory, blue dashed: proposed Layout-Independent allocator, green dashed: baseline and it is not available for two-fin failure because two fin fault-tolerant method is not proposed.}
\label{real}
\end{figure*}

\subsection{Multi layout trajectory tracking}

Simulation experiments were conducted to evaluate the performance of the proposed method in robots with different layouts. The tests were conducted using eeUVsimGazebo \cite{hamamatsu2025icra}. Four different robot layouts with different fin numbers and arrangements were tested. The experiments were conducted using the trajectory tracking controller described in Section 3.A. We quantify the accuracy of the tracking using the Absolute Trajectory Error (ATE) for global consistency and the Relative Trajectory Error (RTE) for local consistency \cite{sturm2012benchmark}. We tested four layouts listed in Table \ref{tab:layout_abcd} and Fig. \ref{sim_fig}. Table \ref{sim_result} shows that the tracking remains stable for all layouts. ATE remains within 0.0937–0.1460 and RTE within 0.0988–0.1755. The average of all layouts (All) is ATE=0.1237 and RTE=0.1281. This indicates that the allocator does not rely on a single fixed geometry and can synthesize a feasible allocation from the provided layout. However, the error patterns differ across layouts. Layout D achieves the best overall performance. This is notable because the model still works with only three fins and does not collapse when redundancy is reduced. 

\subsection{Fault-tolerant scenario performance}

To evaluate the performance of the proposed method in the real-world and the fin failure scenario, experiments were conducted in a swimming pool. The layout of the robot in the pool corresponds to Layout A in simulation, which is excluded from both physics-head pretraining and allocator training, making this a strict zero-shot sim-to-real deployment. Since there is no corresponding baseline for the two-fin failure scenario, we evaluated only the performance of the proposed method. The robot's computer was a Jetson Orin Nano with 8GB RAM. The proposed method and baseline control were tested at 10 Hz. In the fin failure scenario, the failed fin is disabled and set to a state where it does not respond to commands, and control is performed by the remaining operable fins. At this time, the faulty fin is removed from the input graph of the allocator. In all trials, the proposed allocator was applied in a zero-shot manner, using a single trained model without any fine-tuning.
From Table \ref{rw_experiments_tab}, under healthy conditions, Our proposed pipeline achieved ATE=0.200, RTE=0.203, Depth RMSE=0.121, which is almost the same as the layout-specific baseline controller.
With one fin broken, ours maintained similar accuracy and matched the baseline, while showing smaller errors in these metrics. For the two-fin failure scenario, there is no corresponding baseline. The errors increased but the tracking still completed. As shown in Fig. \ref{real}, the trajectory becomes less accurate but still follows the circle without a large deviation. This confirms that, despite some loss of accuracy, the allocator can track in the real world in a zero-shot setting, and in the health and one-failure cases, it performs almost identically to a controller tuned for a specific robot. 

\subsection{Velocity control performance}

\begin{table}[t]
  \centering
  \caption{Real-world experiment result}
  \label{rw_experiments_tab}
  \begin{tabular}{
    l
    S[table-format=1.2] S[table-format=1.2] S[table-format=1.2]
    c c c
    S[table-format=1.2] S[table-format=1.2] S[table-format=1.2]
    S[table-format=1.2] S[table-format=1.2] S[table-format=1.2]
  }
    \toprule
    & \multicolumn{2}{c}{Healthy}
    & \multicolumn{2}{c}{1 fin broken}
    & \multicolumn{2}{c}{2 fins broken} \\
    \cmidrule(lr){2-3}\cmidrule(lr){4-5}\cmidrule(lr){6-7}
    Metrics
    & {Ours} & {Baseline} 
    & {Ours} & {Baseline} 
    & {Ours} & {Baseline*}   \\
    \midrule
    ATE        & 0.1995 & 0.2035 & 0.2169 & 0.2358 & 0.2514 & - \\
    RTE        & 0.2027 & 0.2123 & 0.2145 & 0.2401 & 0.3075 & - \\
    Depth      & 0.1211 & 0.1826 & 0.0964 & 0.2134 & 0.1157 & -  \\ 
    \bottomrule
  \end{tabular}
\begin{tablenotes}
    \item[*] * There are no baseline controller for 2 fin broken scenario.
\end{tablenotes}
\end{table}

\begin{table}[t]
\centering
\caption{Velocity control results}
\label{tab:comparison}
\begin{tabular}{lccc}
\toprule
Metric & Baseline & Ours & Difference (\%) \\
\midrule
Mean Power (W) & $5.45 \pm 0.75$ & $4.80 \pm 1.16$ & $-11.9\%$ \\
Vel. MAE (m/s) & $0.045 \pm 0.014$ & $0.040 \pm 0.018$ & $-11.7\%$ \\
\bottomrule
\end{tabular}
\end{table}

We also evaluated the performance of the proposed method compared to the baseline approach using the same real-world experimental setup. The experiments were conducted five times for 30 seconds each, with target velocities of 0.10, 0.15, and 0.20 m/s. The total power was then calculated as the sum of the inertial power ($P_{\text{inert}} = I \cdot \alpha \cdot \omega$).
Table \ref{tab:comparison} summarizes the statistical results from the experiments. The proposed method demonstrated significant improvements in all key metrics. The mean power consumption decreased by 11.9\%. In terms of control accuracy, the mean absolute error (MAE) of the velocity was reduced by 11.7\%, indicating that the proposed method achieves better velocity tracking while consuming less energy.


\subsection{Discussion and limitation}

There was a sim-to-real tracking accuracy gap between real-world and simulation, and a key cause is that pitch and roll oscillations and intermittent instability are substantially stronger on hardware due to unmodeled disturbances and couplings, which reduces wrench realization quality and degrades tracking; under these conditions our method and the baseline show similar overall performance, suggesting that the dominant limitation is the physical instability and unmodeled dynamics rather than the allocation strategy itself. Our methodology does not require a mathematical inverse model and can perform optimal allocation using only forward data. Therefore, performance could be achieved by using experimental data or high-precision simulation data using CFD for the actuation dataset.

Finally, in terms of computational efficiency, the gradient-based refinement requires iterative forward and backward passes during inference, taking approximately 80 ms per cycle. In our implementation, a 10Hz loop was sufficient, but for agile control, the pipeline needs to be faster. To optimize this, future research could explore model distillation techniques. This allows higher-frequency control or deployment on more resource-constrained hardware.

\section{Conclusion}
In this study, we proposed an allocator for underwater robots that can handle multiple layouts. By training a generalized model using randomly generated layout data, we were able to achieve universal application and zero-shot robot control. The simulation results verified its performance across multiple robot layouts. Similarly, we validated it under real-world fin failure scenarios and compared it with a controller designed for a specific fin layout. The results showed similar performance across different robot layouts and nearly identical performance to that of a controller designed for a specific robot. Future work will extend this model to other types of actuators, such as thrusters and rudders, for even greater versatility.

\section*{Appendix}
The Pretrained model and the training script, including GUI, will be public as open source toolbox. 



\bibliographystyle{IEEEtran}
\bibliography{IEEEabrv,references}

\end{document}